\documentclass[10pt,twocolumn,letterpaper]{article}

\usepackage{cvpr}
\usepackage{times}
\usepackage{epsfig}
\usepackage{graphicx}
\usepackage{amsmath}
\usepackage{amssymb}

\usepackage[pagebackref=true,breaklinks=true,letterpaper=true,colorlinks,bookmarks=false]{hyperref}

\cvprfinalcopy 

\def\cvprPaperID{****} 

\ifcvprfinal\fi
\begin{document}

\title{The Effectiveness of Data Augmentation in Image Classification using Deep Learning}

\author{Jason Wang\\
Stanford University\\
450 Serra Mall\\
{\tt\small zwang01@stanford.edu}
\and
Luis Perez\\
Stanford University\\
450 Serra Mall\\
{\tt\small luis0@stanford.edu}
}

\maketitle

\begin{abstract}
In this paper, we explore and compare multiple solutions to the problem of data augmentation in image classification. Previous work has demonstrated the effectiveness of data augmentation through simple techniques, such as cropping, rotating, and flipping input images. We artificially constrain our access to data to a small subset of the ImageNet dataset, and compare each data augmentation technique in turn. One of the more successful data augmentations strategies is the traditional transformations mentioned above. We also experiment with GANs to generate images of different styles. Finally, we propose a method to allow a neural net to learn augmentations that best improve the classifier, which we call neural augmentation. We discuss the successes and shortcomings of this method on various datasets. 

\end{abstract}

\section{Introduction}


We propose exploring the problem of data augmentation for image and video classification, and evaluating different techniques. It is common knowledge that the more data an ML algorithm has access to, the more effective it can be. Even when the data is of lower quality, algorithms can actually perform better, as long as useful data can be extracted by the model from the original data set. For example, text-to-speech and text-based models have improved significantly due to the release of a trillion-word corpus by Google \cite{EffectivenessOfData}. This result is despite the fact that the data is collected from unfiltered Web pages and contains many errors. With such large and unstructured data sets, however, the task becomes one of finding structure within a sea of unstructured data. However, alternative approaches exist. Rather than starting with an extremely large corpus of unstructured and unlabeled data, can we instead take a small, curated corpus of structured data and augment in a way that increases the performance of models trained on it? This approach has proven effective in multiple problems. Data augmentation guided by expert knowledge \cite{expert_knowledge_augmentation}, more generic image augmentation \cite{general_data_augmentation}, and has shown effective in image classification \cite{classification_augmentation}.\\

The motivation for this problem is both broad and specific. Specialized image and video classification tasks often have insufficient data. This is particularly true in the medical industry, where access to data is heavily protected due to privacy concerns. Important tasks such as classifying cancer types \cite{expert_knowledge_augmentation} are hindered by this lack of data. Techniques have been developed which combine expert domain knowledge with pre-trained models. Similarly, small players in the AI industry often lack access to significant amounts of data. At the end of the day, we've realized a large limiting factor for most projects is access to reliable data, and as such, we explore the effectiveness of distinct data augmentation techniques in image classification tasks.\\

The datasets we examine are the tiny-imagenet-200 data and MNIST \cite{MNISTExample} \cite{MNISTdata} . Tiny-imagenet-200 consists of 100k training, 10k validation, and 10k test images of dimensions 64x64x3. There are a total of 500 images per class with 200 distinct classes. MNIST consists of 60k handwritten digits in the training set and 10k in the test set in gray-scale with 10 classes with image dimensions of 28x28x1. To evaluate the effectiveness of augmentation techniques, we restrict our data to two classes and build constitutional neural net classifiers to correctly guess the class. \\

In particular, we will train our own small net to perform a rudimentary classification. We will then proceed to use typical data augmentation techniques, and retrain our models. Next, we will make use of CycleGAN \cite{cyclegan} to augment our data by transferring styles from images in the dataset to a fixed predetermined image such as Night/Day theme or Winter/Summer. Finally, we explore and propose a different kind of augmentation where we combine neural nets that transfer style and classify so instead of standard augmentation tricks, the neural net learns augmentations that best reduce classification loss. For all the above, we will measure classification performance on the validation dataset as the metric to compare these augmentation strategies. 

\section{Related Work}


This section provides a brief review of past work that has augmented data to improve image classifier performance. The problem with small datasets is that models trained with them do not generalize well data from the validation and test set. Hence, these models suffer from the problem of over-fitting. The reduce overfitting, several methods have been proposed \cite{regularization}. The simplest could be to add a regularization term on the norm of the weights. Another popular techniques are dropout. Dropout works by probabilistically removing an neuron from designated layers during training or by dropping certain connection \cite{dropout1} \cite{dropout2}. Another popular technique is batch normalization, which normalizes layers and allows us to train the normalization weights. Batch normalization can be applied to any layer within the net and hence is very effective \cite{batchnorm}, even when used in generative adversarial networks, such as CycleGAN (\cite{batchnorm_cyclegan}. Finally, transfer learning is a technique in which we take pre-trained weights of a neural net trained on some similar or more comprehensive data and fine tune certain parameters to best solve a more specific problem. \\

Data augmentation is another way we can reduce overfitting on models, where we increase the amount of training data using information only in our training data. The field of data augmentation is not new, and in fact, various data augmentation techniques have been applied to specific problems. The main techniques fall under the category of \textit{data warping}, which is an approach which seeks to directly augment the input data to the model in \textit{data space}. The idea can be traced back to augmentation performed on the MNIST set in \cite{original_image_model}.\\

A very generic and accepted current practice for augmenting image data is to perform geometric and color augmentations, such as reflecting the image, cropping and translating the image, and changing the color palette of the image. All of the transformation are affine transformation of the original image that take the form:
$$
y = Wx + b
$$

The idea has been carried further in \cite{handwriting_success_augmentation}, where an error rate of $0.35\%$ was achieved by generating new training samples using data augmentation techniques at each layer of a deep network. Specifically, digit data was augmented with elastic deformations, in addition to the typical affine transformation. Furthermore, data augmentation has found applicability in areas outside simply creating more data. It has shown to be helpful in generalizing from computer models to real-word tasks.\\

Generative Adversarial Nets (GANs) has been a powerful technique to perform unsupervised generation of new images for training. They have also proven extremely effective in many data generation tasks, such as novel paragraph generation \cite{paragraph_gan}. By using a min-max strategy, one neural net successively generates better counterfeit samples from the original data distribution in order to fool the other net. The other net is then trained to better distinguish the counterfeits. GANs have been used for style transfer such as transferring images in one setting to another setting (CycleGAN). These generated images could be used to train a car to drive in night or in the rain using only data collected on sunny days for instance. Furthermore, GANs have been effective even with relatively small sets of data \cite{gan_small_data} by performing transfer learning techniques. Additionally, they have shown to be extremely good at augmenting data sets, such as increasing the resolution of input images \cite{gan_augment}. \\

Finally, we explore methods where we train the neural net to both augment and classify simultaneously. A similar approach was tried in \cite{smart_augmentation}, though the approach there learned different weights for combining already existing techniques. In our case, we can train a style transfer network to learn how to best generate data augmentations. The goal is to not only reduce over-fitting via augmentation but also to augment data in a way such that to best improve the classifier. These methods do not necessarily generate images that resemble the training set as techniques like affine transformation or GANs would. Therefore, it saves the effort of needing manual transformations or correlations between the generated images with a method like GANs and the original image. 

\section{Methods}

We propose two different approaches to data augmentation. The first approach is generate augmented data before training the classifier. For instance, we will apply GANs and basic transformations to create a larger dataset. All images are fed into the net at training time and at test time, only the original images are used to validate. The second approach attempts to learn augmentation through a pre-pended neural net. At training time, this neural net takes in two random images from the training set and outputs a single "image" so that this image matches either in style or in context with a given image from the training set. This output, which represents an augmented image produced by the network, is fed into the second classifying network along with the original training data. The training loss is then back-propagated to train the augmenting layers of the network as well as the classification layers of the network. In test time, images from the validation or test set is ran through only the classification network. The motivation is to train a model to identify the best augmentations for a given dataset. The remainder of this section will go into detail of the data augmentation tricks we tried. \\

\subsection{Traditional Transformations}

Traditional transformations consist of using a combination of affine transformations to manipulate the training data \cite{smiles}. For each input image, we generate a "duplicate" image that is shifted, zoomed in/out, rotated, flipped, distorted, or shaded with a hue. Both image and duplicate are fed into the neural net. For a dataset of size $N$, we generate a dataset of $2N$ size. 

\begin{center}
\includegraphics[scale=0.4]{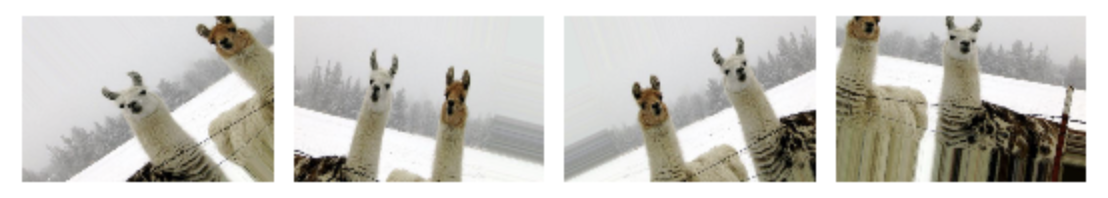}
Figure I: Traditional Transformations
\end{center}

\subsection{Generative Adversarial Networks}

For each input image, we select a style image from a subset of 6 different styles: Cezanne, Enhance, Monet, Ukiyoe, Van Gogh and Winter. A styled transformation of the original image is generated. Both original and styled image are fed to train the net. More detail about the GANs and style transfer can be viewed on the cited paper \cite{cyclegan}. 

\begin{center}
\includegraphics[scale=0.52]{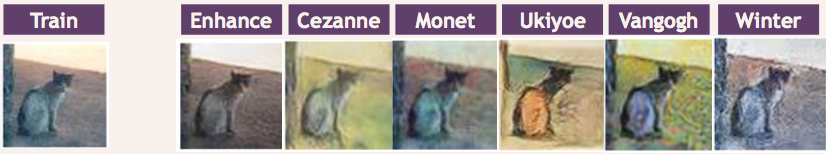}
Figure II: Style Transformations via GANs
\end{center}

\subsection{Learning the Augmentation}

During the training phase, there are two parts to the network. The augmentation network takes in two images from the same class as the input image and returns a layer the same size as a single image. This layer is treated as an "augmented" image. The augmented image as well as the original input image are then passed into the second network, the classification network. The classification loss at the end of the network is a cross entropy loss on the sigmoids of the scores of classes. An addition loss is computed at the end of the augmentation network to regulate how similar the augmented image should be to the input image. The overall loss is a weighted sum of these two losses. We try three different approaches: 
\begin{enumerate}
\item Content loss 
\item Style loss via gram matrix 
\item No loss is computed at this layer
\end{enumerate}

More details about the architecture of the layers will be described in the Experiments section. We implement a small 5-layer CNN to perform augmentation. The classifier is a small 3-layer net with batch normalization and pooling followed by 2 fully connected layers with dropout. This is much similar to VGG16 in structure but smaller in interest of faster training for evaluation. We aren't aiming for the best classifier. We are exploring how augmentation tricks improve classification accuracy, reduce over-fitting, and help the networks converge faster. 

\section{Datasets and Features}

There are three sets of images that we experimented with. Each dataset is a small dataset with two classes. A small portion of the data is held aside for testing. The remaining images are divided by a 80:20 split between training and validation. \\

Our first data set is taken from tiny-imagenet-200. We take 500 images from dogs and 500 images from cats. 400 images for each class is allocated to the training set. The remaining 100 in each class forms the validation set. The images are 64x64x3. RGB values are also normalized for each color in the preprocessing step. \\

The second data set is also taken from tiny-imagenet-200 except we replace cats with goldfish. The reason for this change is that goldfish look very different from dogs whereas cats visually are very similar. Hence CNNs tend to have a harder time distinguishing cats. Finally, cats and dogs have similar styles whereas images from the goldfish tend to have very bright orange styles. \\

Lastly, the final dataset is 2k images from MNIST, 1000 from each class. We perform the task of distinguishing 0's from 8's. MNIST images are 28x28x1 and are in gray scale. Again, images are normalized in the preprocessing step. MNIST is much more structured than imagenet so that digits are always centered. The motivation is that MNIST provides a very simple dataset with simple images. Are patterns in the more complex images also observed in simpler images? 

\section{Experiments}



To test the effectiveness of various augmentation, we run 10 experiments on the image-net data. The results of the experiments are tabulated in the following table. All experiments are run for 40 epochs at the learning rate of 0.0001 using Adam Optimization. The highest test accuracy at all the epochs is reported as the best score. \\

Once we obtained the augmented images, we feed them into a neural net that does classification. We name this neural net the SmallNet since it only has 3 convolution layers paired with a batch normalization and max pool layer followed by 2 fully connected layers. The output is a score matrix for the weights for each class. The layers of the network is detailed below although the specific net is not very important. Any net that can reliably predict the classes suffices. Hence, one can replace this net with VGG16 with fine-tuning on the fully connected and last convolution layers to allow for sufficient training. \\

\begin{center}
\underline{\textbf{SmallNet}}

\noindent 1. Conv with 16 channels and 3x3 filters. Relu activations. \\
2. Batch normalization. \\
3. Max pooling with 2x2 filters and 2x2 stride. \\
4. Conv with 32 channels and 3x3 filters. Relu activations. \\
5. Conv with 32 channels and 3x3 filters. Relu activations. \\
6. Batch normalization. \\
7. Max pooling with 2x2 filters and 2x2 stride. \\
8. Fully connected with output dimension 1024. Dropout. \\
9. Fully connected layer with output dimension 2. \\
\end{center}

Augmenting data via a neural net is achieved by concatenating two images of the same class to create an input of 6 channels deep (2 if gray scale). The goal of this layer is to use a CNN to generate an image of the same height and width of the input with 3 channels deep. We can also add an additional loss term at the end of this layer that compares the output of the augmented layers to a third image from the same class. In this arrangement, the augmented layer generates outputs that are similar to the third image, which acts as a regularizer. Regardless, the motivation behind this idea that we can use paired data from a small data set to create new "images" to perform better training. A dataset of size $N$ can create $N^2$ pairs, a magnitude increase. The architecture of the augmentation network is detailed below. 

\begin{center}
\underline{\textbf{Augmentation Network}}

\noindent 1. Conv with 16 channels and 3x3 filters. Relu activations. \\
2. Conv with 16 channels and 3x3 filters. Relu activations. \\
3. Conv with 16 channels and 3x3 filters. Relu activations. \\
4. Conv with 16 channels and 3x3 filters. Relu activations. \\
5. Conv with 3 channels and 3x3 filters. \\
\end{center}

At training time, we generate a batch of images called the training batch. This image is fed into SmallNet and gradients are back-propagated to help improve SmallNet. Subsequently, pairs of images are randomly sampled from the same class and fed into AugNet to generate an augmented image which is passed into SmallNet. The weights of both neural nets are updated. At test time, images are fed into SmallNet, which does all the work to classify the image. 

\begin{center}
\includegraphics[scale=0.41]{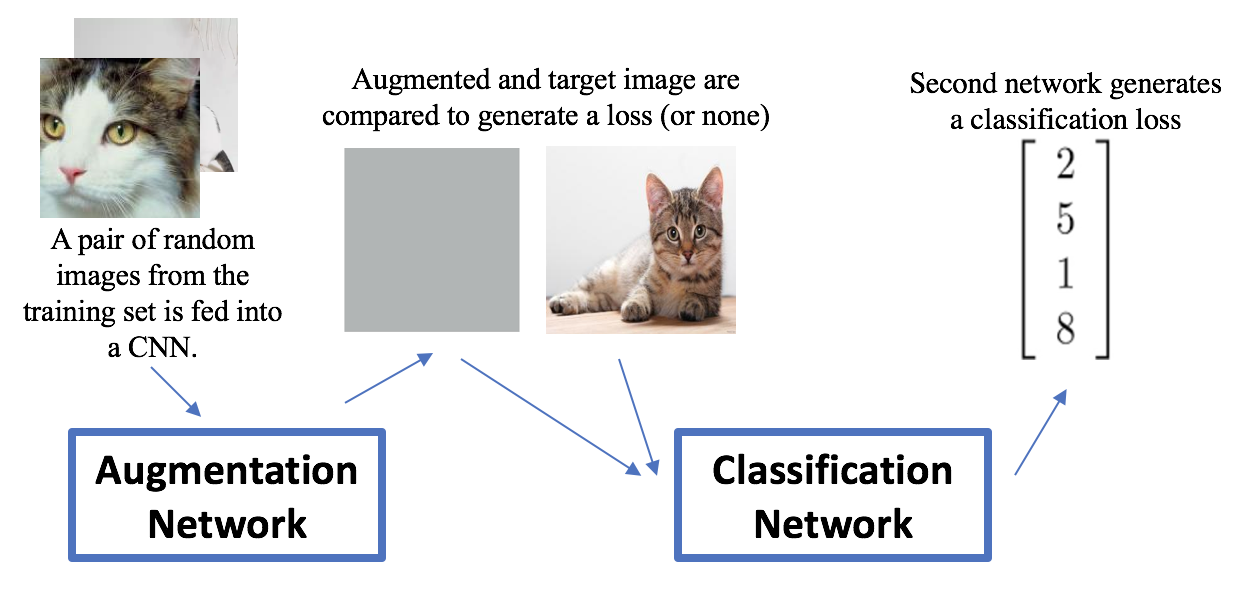}
Figure III: Training model 
\end{center}

\begin{center}
\includegraphics[scale=0.48]{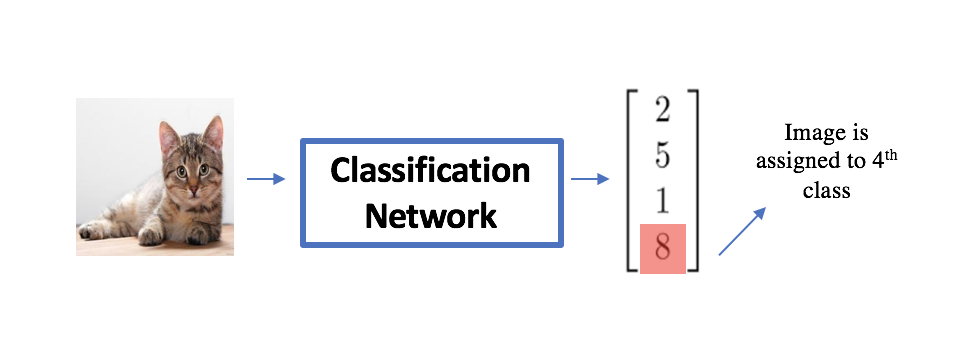}
Figure IV: Testing/Validation model 
\end{center}

To determine the loss, we wanted a combination of a classification loss, $L_c$, and an augmentation loss, $L_a$. There are two augmentation losses we considered. The content loss is the mean squared error between the augmented image $A$ and the target image $T$, where $D$ is the length of images $A$ and $T$. 
\[ L_a^{content} = \frac{1}{D^2} \sum_{i,j}(A_{ij} - T_{ij})\]

The style loss is a content loss on the gram matrix of the augmented image $A$ and target image $T$. The gram matrix of feature map $F$ is defined below. We apply the gram matrix on the raw images. 

\[G_{ij}  = \sum_k F_{ik} F_{jk}\]

Then the loss is defined below where $C$ is the number of channels. 

\[L_a^{style} = \frac{1}{C^2} \sum_{i,j}(G_{ij}^A - G_{ij}^T)\]

Finally, we consider the case where no loss is computed. Classification loss is a multi-class cross entropy loss on the sigmoids of the scores produced by SmallNet. The final loss is weighted sum of the two losses. Note that setting $\beta =0$ is equivalent to having no augmentation loss. 

\[ \alpha L_c + \beta L_a\]

Models are trained with learning rate 0.0001 with ADAM optimization. Tensorflow on GPU was used to train the models although they train reasonable fast on CPU as well (around 30 seconds per epoch). 

\section{Results}

The first set of experiments involve classifying dogs vs cats. Several experiments were done to explore the effectiveness of augmentation techniques. \\

\noindent 1. EXPERIMENTS ON TRADITIONAL TRANSLATION

After manually performing traditional augmentation on images from each class, we train SmallNet on all images. \\

\noindent 2. EXPERIMENTS ON GANS 

Randomly select a style to generate via GANs for each image and train SmallNet. \\

\noindent 3. EXPERIMENTS ON  NEURAL NET AUGMENTATION 

For each image, we select two random images from the same class and concatenate them so that the first 3 channels are image 1 and the last 3 channels are image 2. This is inputted into the Augmentation net which returns a augmented image of the size 64x64x3. Another random image is selected and an augmentation loss is computed. We explore content and style loss, explained in the Experiments section, between the randomly selected image and the output of the Augmentation net. Both images are separately used to train SmallNet and a classification loss is computed. The overall loss is a linear combination of the two losses. We found that the exact ratio didn't matter, but we selected $\beta = 0.25$ weight for augmentation loss and $\alpha = 0.75$ for classification loss. We also explored not using any augmentation loss at all. \\

\noindent 4. SAME EXPERIMENTS ON DOG VS GOLDFISH 

The 3 experiments above were replicated on goldfish vs dog classfication. This classification problem is an easier one than dogs and cats since goldfish typically look distinctively different with lots of orange color. We want to see if the augmentation strategies are robust to different data. \\

\noindent 5. CONTROL EXPERIMENT 

A possible explanation for any improved validation is that we used a more complicated net, which typically prevents overfitting and allows for finer training. To control against this, we input the same two images into Augmentation Net. The result of the net and the input image are both used to train SmallNet. The effect is a proxy to training a large 10-layer net on a single image without augmentation (pairing of different images). The control experiment is performed on both dogs vs cats and dogs vs fish. \\

\noindent 6. EXPERIMENTS WITH MNIST 

We train MNIST data to explore neural augmentation strategies and look qualitatively at how data is augmented. These images should provide us an insight into how neural augmentation works with structured data since digits are centered and relatively the same size. Imagenet data is unstructured and the key features of each data image appears in many orientation and positions. Hence, loss functions such as style loss and content loss tend to produce "mean" or "average" images that are duller and noisy. Finally because we're using greyscale data, the gram matrix wouldn't produce any meaning so we only experiment with the effects of content loss. \\

The results of the experiments are tabulated below. The best validation accuracy in 40 epochs is recorded. Table 1 shows all experiments on dog vs goldfish data. Table 2 shows the results of experiments on dog vs cat data. Table 3 explores neural augmentation on MNIST data. \\

\begin{center}
\begin{tabular}{ |p{4cm}||p{2cm}|  }
 \hline
 \multicolumn{2}{|c|}{Dogs vs Goldfish} \\
 \hline
 Augmentation & Val. Acc.\\
 \hline
None & 0.855 \\
Traditional   & 0.890  \\
GANs   & 0.865 \\
Neural + No Loss  & \underline{\textbf{0.915}}   \\
Neural + Content Loss & \underline{0.900}  \\
Neural + Style & \underline{0.890}  \\
Control & 0.840 \\
 \hline
\end{tabular}\\
Table I: Quantitative Results on Dogs vs Goldfish
\end{center}

\begin{center}
\begin{tabular}{ |p{4cm}||p{2cm}|  }
 \hline
 \multicolumn{2}{|c|}{Dogs vs Cat} \\
 \hline
 Augmentation & Val. Acc.\\
 \hline
None & 0.705 \\
Traditional   & \textbf{0.775}  \\
GANs   & 0.720 \\
Neural + No Loss  & \underline{0.765}    \\
Neural + Content Loss & \underline{0.770}  \\
Neural + Style & \underline{0.740}  \\
Control & 0.710 \\
 \hline
\end{tabular}\\
Table II: Quantitative Results on Dogs vs Cats
\end{center}

\begin{center}
\begin{tabular}{ |p{4cm}||p{2cm}|  }
 \hline
 \multicolumn{2}{|c|}{MNIST 0's and 8's} \\
 \hline
 Augmentation & Val. Acc.\\
 \hline
None & 0.972 \\
Neural + No Loss  & \textbf{\underline{0.975}}    \\
Neural + Content Loss & \underline{0.968}  \\
 \hline
\end{tabular}\\
Table III: MNIST
\end{center}

Neural augmentation performs remarkably better than no augmentation. In the dogs vs cats problem, the neural augmentation performs the best with a 91.5\% to 85.5\% compared to no augmentation. In the dogs vs fish problem, the neural augmentation performed the second best with 77.0\% to 70.5\%. While the traditional augmentation performs almost as well at a smaller time expense, this doesn't preclude us from combining different augmentation strategies. A strategy that first performs traditional augmentations, then pairs up data for neural augmentation could potentially beat all experiments we tested. \\

Only the control (see experiment 5) does worse than no augmentation. Perhaps we are dealing with a larger net and are using the wrong learning rate. This hypothesis is supported by the inability for the net to converge on the training data (loss doesn't decrease to zero and/or inability to perfectly classify the training data). The lack of improvement provides evidence that adding layers for a classification data with little data doesn't reduce overfitting or help the model generalize better. \\

We also note that neural nets with various augmentation loss (or none) perform relatively the same. In fact, during training, the content loss and style loss didn't decrease at all. Content loss decreased from about 1.6 to 1.3-1.5 after 30 epochs and never converged. Style loss hovered around 0.5. Assuming that we can actually minimize these losses is equivalent to claiming that we could create an oracle to perfectly recreate images of any dog from any pair of images of dogs. This is an outlandish task given just convolutional layers. That is not to say these losses are completely useless. They act like regularization terms, ensuring that the Augmentation net doesn't generate images so different from data in the training set. \\

Neural augmentation has no effect on the MNIST dataset. We hypothesis that a simple CNN already performs so well on MNIST so that neural augmentation provides no benefits. We also hypothesis that the digits are already so simple that combining features don't really add any additional information. Finally there are just so many variations in the digits so augmented data don't provide "new" images that the network has not seen before.\\

Some of the images generated by neural augmentation are quite remarkable. In most cases, the augmented images are a combination of the source images. The neural picks out the golden bodies of the fish and merges them together. For instance, in sample II, the neural augmentation picks out the large goldfish in the second source image and the orange fish on the left in the first source image while smoothing out the background grass.  

\begin{center}
\includegraphics[scale=0.33]{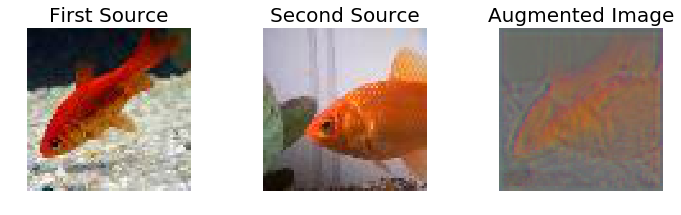}
Figure V: Goldfish sample I
\end{center}

\begin{center}
\includegraphics[scale=0.33]{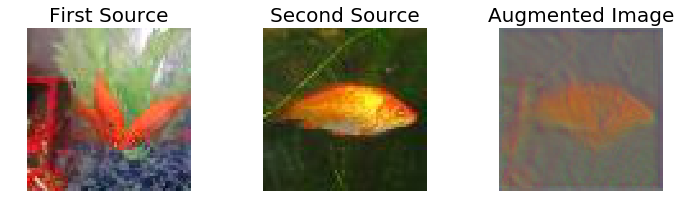}
Figure VI: Goldfish sample II
\end{center}

The sampling for dogs is quite similar. The augmented image picks up the characteristics of the second source image while preserving only the contours of the other dog. Features such as the contours of the nose and the legs of the other dog are somewhat visible. 

\begin{center}
\includegraphics[scale=0.33]{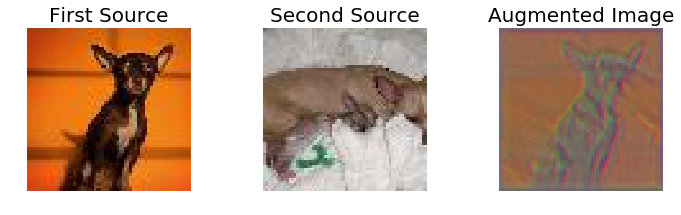}
Figure VII: Dog sample I
\end{center}

However not all augmented images have any visual meaning. Even though no defining shape of dogs are created in the following two images, the augmented images are a bunch of contours of ears and legs which are defining characteristics of the dog. 

\begin{center}
\includegraphics[scale=0.33]{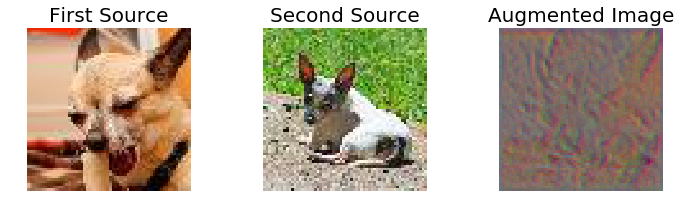}
Figure VIII: Dog sample II
\end{center}

\begin{center}
\includegraphics[scale=0.33]{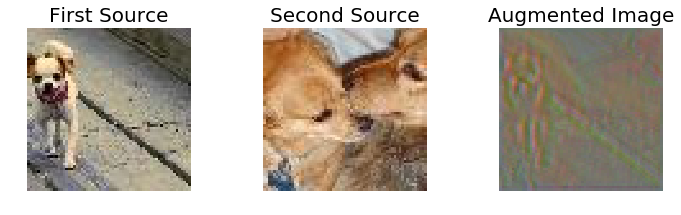}
Figure IX: Dog sample III
\end{center}

In all cases, it seems that the generated images that best improve performance have some form of regularization so that the color is faded and the background noise is faded out. Regarding colors, orange background colors in dog images are always picked up in the background to contrast images of goldfish. The next figure shows this property where the yellow wall paper is transformed into an orange hue. It's really fascinating that the despite the dogs' ears matching the tones of the background in both source images, the augmented image picks up on those details and colors them greenish in contrast to the orange background. 

\begin{center}
\includegraphics[scale=0.33]{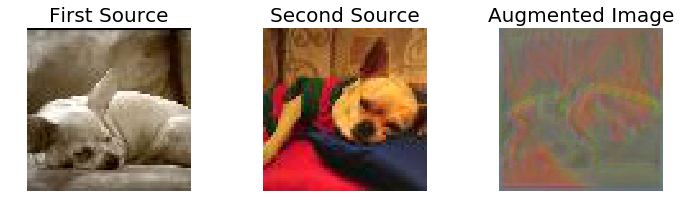}
Figure X: Dog sample IV
\end{center}

The accuracy plot at each epoch shows that neural augmentation helps a little in preventing overfitting. For most of the first 20 epochs of training, the training accuracy with augmentation is slightly lower than the training accuracy without augmentation. This implies that learning augmentation helps with generalizing the classifier. A comparison of the losses would be more apt but not viable in this case since the various experiments use different loss functions. 

\begin{center}
\includegraphics[scale=0.33]{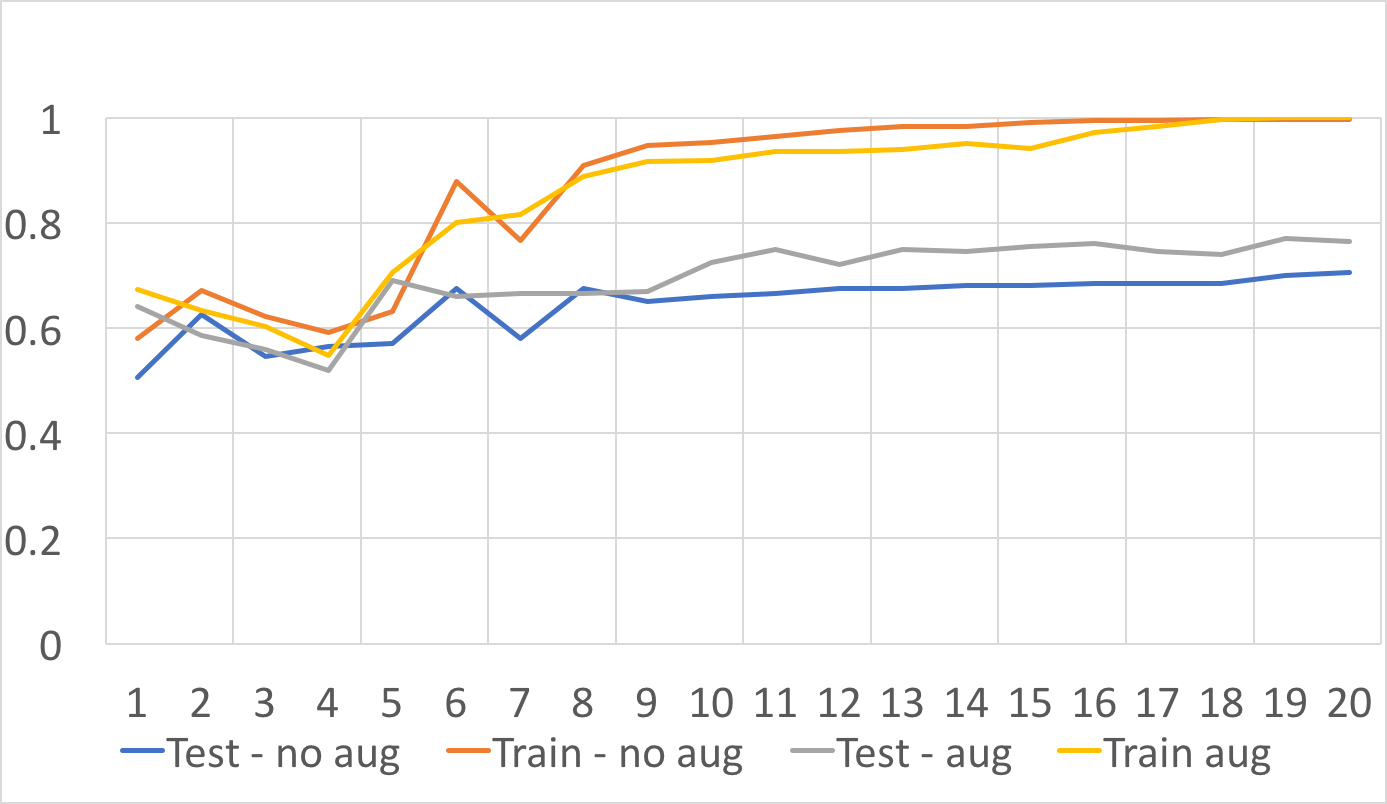}
Figure XI: Accuracy plots
\end{center}

\section{Conclusion/Future Work}

Data augmentation has been shown to produce promising ways to increase the accuracy of classification tasks. While traditional augmentation is very effective alone, other techniques enabled by CycleGAN and other similar networks are promising. We experimented with our own way of combining training images allowing a neural net to learn augmentations that best improve the ability to correctly classify images. If given more time, we would like to explore more complex architecture and more varied datasets. To mimic industrial applications, using a VGG16 instead of SmallNet can help us determine if augmentation techniques are still helpful given complex enough networks that already deal with many overfitting and regularization problems. Finally, although GANs and neural augmentations do not perform much better than traditional augmentations and consume almost 3x the compute time or more, we can always combine data augmentation techniques. Perhaps a combination of traditional augmentation followed by neural augmentation further improves classification strength.  \\

Given the plethora of data, we would expect that such data augmentation techniques might be used to benefit not only classification tasks lacking sufficient data, but also help improve the current state of the art algorithms for classification. Furthermore, the work can be applicable in more generic ways, as "style" transfer can be used to augment data in situations were the available data set is unbalanced. For example, it would be interesting to see if reinforcement learning techniques could benefit from similar data augmentation approaches. We would also like to explore the applicability of this technique to videos. Specifically, it is a well known challenge to collect video data in different conditions (night, rain, fog) which can be used to train self-driving vehicles. However, these are the exact situations under which safety is the most critical. Can our style transfer method be applied to daytime videos so we can generate night time driving conditions? Can this improve safety? If such methods are successful, then we can greatly reduce the difficulty of collecting sufficient data and replace them with augmentation techniques, which by comparison are much more simpler. 


{\small
\bibliographystyle{ieee}
\bibliography{egbib}
}

\end{document}